\documentclass[]{article}
\usepackage[letterpaper]{geometry}
\usepackage{mtsummit2017}
\usepackage{times}
\usepackage{natbib}
\usepackage{layout}

\usepackage{amssymb,amsmath,bm}
\usepackage{textcomp}
\usepackage{colortbl}
\usepackage{url}
\usepackage{multirow}
\usepackage{latexsym}
\usepackage{graphicx}
\usepackage[english, french]{babel}
\usepackage[T1]{fontenc}
\usepackage[utf8]{inputenc} 
\usepackage{algpseudocode} 
\usepackage{algorithm}
\usepackage{algorithmicx}
\usepackage[table, xcdraw]{xcolor}
\usepackage{enumitem}
\usepackage{pdfpages}
\usepackage{arydshln}
\usepackage{siunitx} 

\usepackage[labelsep=period]{caption} 
\AtBeginDocument{} 

\begin{document}
\shorthandoff{:} 
\NoAutoSpacing 

\title{\bf Disentangling ASR and MT Errors \\ in Speech Translation}

\author{\name{\bf Ngoc-Tien Le} \hfill  \addr{ngoc-tien.le@univ-grenoble-alpes.fr}
\AND
        \name{\bf Benjamin Lecouteux} \hfill \addr{benjamin.lecouteux@univ-grenoble-alpes.fr}
\AND
\name{\bf Laurent Besacier} \hfill \addr{laurent.besacier@univ-grenoble-alpes.fr}
\AND         
        \addr{University Grenoble Alpes, CNRS, Grenoble INP, LIG, F-38000 Grenoble, France}
}




\maketitle
\pagestyle{empty}


%
\begin{abstract}

The main aim of this paper is to investigate automatic quality assessment for spoken language translation (SLT). More precisely, we investigate SLT errors that can be due to transcription (ASR) or to translation (MT) modules. This paper investigates automatic detection of SLT errors using a single classifier based on joint ASR and MT features. We evaluate both 2-class ($good$/$bad$) and 3-class ($good$/$bad_{ASR}$/$bad_{MT}$) labeling tasks. The 3-class problem necessitates to disentangle ASR and MT errors in the speech translation output and we propose two label extraction methods for this non trivial step. This enables - as a by-product - qualitative analysis on the SLT errors and their origin (are they due to transcription or to translation step?) on our large in-house corpus for French-to-English speech translation. 



\end{abstract}


\section{Introduction}

\label{sec:intro}

This paper addresses a relatively new quality assessment task: error detection in spoken language translation (SLT) using both automatic speech recognition (ASR) features and machine translation (MT) features. To our knowledge, the first attempts to design error detection for speech translation, using both ASR and MT features, are our own work  
\citep{besacier14,besacier-asru-2015}
 which is further extended in this paper submission.

\textbf{Contributions}  (1) This paper extends previous work 
\citep{besacier14,besacier-asru-2015}
 in 2-class ($good$/$bad$) error detection in SLT using a single classifier based on joint ASR and MT features (2) in order to disentangle ASR and MT errors in SLT, we extend error detection to a 3-class problem ($good$/$bad_{ASR}$/$bad_{MT}$) where we try to find the source of the SLT errors (3) two methods are compared for setting such 3-class labels on our corpus and a first attempt to automatically detect errors and their origin in a SLT output is presented at the end of this paper.

\textbf{Outline} The outline of this paper goes simply as follows: Section 2 formalizes error detection in SLT and presents our experimental setup. Section 3 proposes two methods to disentangle ASR and MT errors in SLT output and presents statistics on 
a large French-English corpus.
Section 4 presents our 2-class and 3-class error detection results while section 5 concludes this work and gives some perspectives.

\section{Automatic Error Detection in Speech Translation}
\label{sec:WCE}

\subsection{Formalization}
A quality estimation (or error detection) component in speech translation solves the equation:

%

\begin{equation}
\label{eq:wce_component}
\hat{q}=\underset{q}{\operatorname{argmax} } \{p_{SLT}(q|x_{f},f, \hat{e}) \}
\end{equation}


where $x_{f}$ is the given signal in the source language; $\hat{e}\footnote{written simply $e$ for convenience in any other equations} = (e_{1}, e_{2},...,e_{N})$  is the most probable target language sequence from the spoken language translation (SLT) process; $f = (f_{1}, f_{2},...,f_{M})$ is the transcription of $x_{f}$; $q = (q_{1}, q_{2},...,q_{N})$ is a sequence of error labels on the target language and $q_{i}\in\{good,bad\}$\footnote{ at this point $q_{i}$ takes two values (G/B) but will evolve to 3 labels later on in section  3}. This is a sequence labeling task that can be solved with several machine learning techniques such as Conditional Random Fields (CRF) \citep{lafferty01}. However, for that, we need a large amount of training data for which a quadruplet $(x_{f},f,e,q)$ is available.

As it is much easier to obtain data containing either the triplet $(x_{f},f,q)$ 
(ASR output + manual references and  error labels inferred from WER)
or the triplet $(f,e,q)$ 
(MT output + manual post-editions
and  error labels inferred using tools such as TERp-A \citep{snover08}) we can also recast 
error detection with the following equation:

\begin{equation}
\label{eq:wce_asr_and_mt}
\hat{q}=\underset{q}{\operatorname{argmax} } \{p_{ASR}(q|x_{f},f)^\alpha*p_{MT}(q|e,f)^{1-\alpha}\}
\end{equation}
where $\alpha$ is a weight giving more or less importance to error detector on transcription compared to error detector on translation.



\subsection{Dataset, ASR and MT Modules}
\label{subsec:dataset}

\subsubsection{Dataset}
In this paper, we use our in-house corpus made available on a \textit{github} 
repository 
\footnote{\url{https://github.com/besacier/WCE-SLT-LIG/}} 
for reproductibility. 
The \textit{dev} set and \textit{tst} set of this corpus were recorded by french native speakers.
Each sentence was uttered by 3 speakers, leading to 2643 and 4050 speech recordings for \textit{dev} set and \textit{tst} set, respectively. 
For each speech utterance, a 
quintuplet containing: ASR output ($f_{hyp}$), verbatim transcript ($f_{ref}$), 
text translation output ($e_{hyp_{mt}}$), speech translation output 
($e_{hyp_{slt}}$) and post-edition of translation ($e_{ref}$) is  
available. 
The total length of the union of \textit{dev} and \textit{tst} is 16h52 
(42 speakers - 5h51 for \textit{dev}  and 11h01 for \textit{tst}).



\subsubsection{ASR Systems}
\label{subsec:asr_system}

To obtain the speech transcripts ($f_{hyp}$), we built a French ASR system based on KALDI toolkit \citep{Povey_ASRU2011}. 
Acoustic models are trained using several corpora (ESTER, REPERE, ETAPE and BREF120) representing more than 600 hours of french transcribed speech. 
We use two 3-gram language models trained on French ESTER corpus \citep{Galliano06corpusdescription} as well as on French Gigaword (vocabulary size are respectively 62k and 95k). ASR systems LM weight parameters are tuned through WER on  \textit{dev} corpus. 
\textit{Table \ref{tab:mt-slt-res-dev-tst-set}} presents the performances obtained by both ASR systems.


\subsubsection{SMT System}
\label{subsec:smt_system}
We used \textit{moses} phrase-based translation toolkit \citep{koehn07} to translate French ASR into English ($e_{hyp}$).
This medium-size system was trained using a subset of data provided for IWSLT 2012 evaluation  \citep{iwslt2012_campaign}: Europarl, Ted and News-Commentary corpora.
The total amount is about 60M words. 
We used an adapted target language model trained on specific data (News Crawled 
corpora) similar to our evaluation corpus (see 
\citep{potet10}). 


\subsection{Obtaining Error Labels for SLT}
\label{subsec:obtaining_label}


After building an ASR system, we  have a new element of our desired quintuplet: the ASR output $f_{hyp}$. It is the noisy version of our already available verbatim transcripts called $f_{ref}$. 
This ASR output ($f_{hyp}$) is then translated by the SMT system 
\citep{potet10} 
already mentioned in subsection \ref{subsec:smt_system}.
This new output translation is called $e_{hyp_{slt}}$ and it is a degraded version of $e_{hyp_{mt}}$ (translation of $f_{ref}$). 
To infer the quality (G, B) labels of our speech translation output $e_{hyp_{slt}}$, we use TERp-A toolkit \citep{snover08} between $e_{hyp_{slt}}$ and $e_{ref}$ (more details can be found in our former paper 
\citep{besacier-asru-2015}).
Table \ref{tab:mt-slt-res-dev-tst-set} 
summarizes baseline ASR, MT and SLT performances obtained on our corpora, as well as the distribution of $good$ (G) and $bad$ (B) labels inferred for both tasks. Logically, the percentage of (B) labels increases from MT to SLT task in the same conditions and it decreases when ASR system improves.

%


\begin{table}[ht]
\centering
\begin{tabular}{lllllllll}
\noalign{\smallskip}\hline
\bf Task & \multicolumn{2}{c}{\textit{\textbf{ASR (WER)}}}  & \multicolumn{2}{c}{\textit{\textbf{MT (BLEU)}}} & \multicolumn{2}{c}{\textit{\textbf{\% G (good))}}} & \multicolumn{2}{c}{\textit{\textbf{\% B (bad)}}}\\
\noalign{\smallskip}\hline
 & \multicolumn{1}{c}{\textit{dev} set} & \multicolumn{1}{c}{\textit{tst} set} & \multicolumn{1}{c}{\textit{dev} set} & \multicolumn{1}{c}{\textit{tst} set} & \multicolumn{1}{c}{\textit{dev} set} & \multicolumn{1}{c}{\textit{tst} set} & \multicolumn{1}{c}{\textit{dev} set} & \multicolumn{1}{c}{\textit{tst} set} \\
\noalign{\smallskip}\hline
MT & \- & \-  &  49.13\% & 57.87\% & 76.93\% & 81.58\%  &  23.07\%  & 18.42\% \\
\noalign{\smallskip}\hline
SLT (ASR1) & 21.86\% & 17.37\% & 26.73\%  & 36.21\% & 62.03\% & 70.59\% & 37.97\%  & 29.41\% \\
\noalign{\smallskip}\hline
SLT (ASR2) & 16.90\% & 12.50\% & 28.89\%  & 38.97\% & 63.87\% & 72.61\% & 36.13\%  & 27.39\%\\
\noalign{\smallskip}\hline
\end{tabular}
\caption{\label{tab:mt-slt-res-dev-tst-set}ASR, MT and SLT performances on our \textit{dev} set and \textit{tst} set.}
\end{table}


%
%


\section{Disentangling ASR and MT Errors}
\label{sec:disentangle}

In previous section, we only extract $good/bad$ labels from the SLT output while it might be interesting to move from a 2-class problem to a 3-class problem in order to label our SLT hypotheses with one of the 3 following labels: $good$ ($G$), \textit{asr-error} (\textit{B\_ASR}) and \textit{mt-error} (\textit{B\_MT}).  Before training automatic systems for error detection, we need to set such 3-class labels on our $dev$ and $test$ corpora. For that, we propose, in the next sub-sections, two slightly different methods to extract them. 
The first one is based on word alignments between SLT and MT and
the second one is based on a simpler SLT-MT error subtraction.

\subsection{Method 1 - Word Alignments between MT and SLT}
In machine translation, fertility of a source word designs to how many output words  it  translates. If we transpose this definition to our disentangling problem, then \emph{fertility of an MT error} designs how many erroneous words - in the SLT output - it is aligned to.
From this simple definition, we derive our first way (\textit{Method 1}) to generate 3-class annotations.



Let $\hat{e}_{slt} = (e_1, e_2, …, e_n)$: the set of SLT hypotheses ($e_{hyp_{slt}}$); $e_{k_{j}}$ denotes the  $j^{th}$ word in the sentence $e_k$, where $1\leq k \leq n$

Let $\hat{e}_{mt} = (e'_1, e'_2, …, e'_n)$: the set of MT hypotheses ($e_{hyp_{mt}}$); $e'_{k_{i}}$ denotes the $i^{th}$ word in the sentence $e'_k$, where $1\leq k \leq n$

Let $L = (l_1, l_2, …, l_n)$: the set of the word alignments from sentences in $e_{hyp_{slt}}$ to related sentences in $e_{hyp_{mt}}$, where $l_{k}$ contains the word alignments from sentence $e_{k}$ to relevant sentence $e'_{k}$, $1\leq k \leq n$; $(e_{k_{j}}, e'_{k_{i}})$ = \textit{True}, if there is one word alignment between $e_{k_{j}}$ and $e'_{k_{i}}$; $(e_{k_{j}}, e'_{k_{i}})$ = \textit{False}, otherwise.

Our algorithm for \textit{Method 1} is defined as \textit{Algorithm 1}. This method relies on word alignments and uses MT labels. We also propose a simpler method in the next section.

\begin{algorithm}[!ht]
\caption{\label{algo:method1_from_SLT_to_MT}\textit{Method 1} - Using word alignments between MT and SLT }
\begin{algorithmic}
\State $list\_labels\_result\gets empty\_list$

\For{each sentence $e_k$ $\in$ $\hat{e}_{slt}$}

	\State $list\_labels\_sent\gets empty\_list$
	
	\For{$j\gets 1$ to $NumberOfWords(e_k)$}
	
		\If {$label(e_{k_{j}}) = $ `G'}
		    \State add `G' to $list\_labels\_sent$
		\ElsIf {Existed Word Alignment $(e_{k_{j}}, e'_{k_{i}})$ and $label(e'_{k_{i}})$=`B'}
			\State add `$B\_MT$' to $list\_labels\_sent$
		\Else
			\State add `$B\_ASR$' to $list\_labels\_sent$
		\EndIf		
	\EndFor

	\State add $list\_labels\_sent$ to $list\_labels\_result$
\EndFor

\end{algorithmic}
\end{algorithm}



\subsection{Method 2 - Subtraction between SLT and MT Errors}

Our second way to extract 3-class labels (\textit{Method 2}) focuses on the differences between SLT hypothesis ($e_{hyp_{slt}}$) and MT hypothesis ($e_{hyp_{mt}}$). We call it \textit{subtraction between SLT and MT errors} because we simply consider that errors present in SLT and not present in MT are due to ASR. This method has a main difference with the previous one: it does not rely on the extracted labels for MT.

Our intuition is that the number of \textit{mt-error}s estimated will be slightly lower than for \textit{Method 1} since we first estimate the number of \textit{asr-error}s and the rest is considered - by default - as \textit{mt-error}s.  



%

%
%

With the same notations of \textit{Method 1}, but highlighting that $L = (l_1, l_2, …, l_n)$ is the set of alignments through edit distance between $e_{hyp_{slt}}$ and $e_{hyp_{mt}}$, where $l_{k_i}$ corresponds to ``Insertion'', ``Substitution'', ``Deletion'' or ``Exact''. 
Our algorithm for \textit{Method 2} is defined as follows.

\begin{algorithm}[!ht]
\caption{\label{algo:method2_from_SLT_to_MT}\textit{Method 2} - Subtraction between SLT and MT errors}
\begin{algorithmic}
\State $list\_labels\_result\gets empty\_list$

\For{each sentence $e_k$ $\in$ $\hat{e}_{slt}$}

	\State $list\_labels\_sent\gets empty\_list$
	
	\For{$j\gets 1$ to $NumberOfWords(e_k)$}
	
		\If {$label(e_{k_{j}}) = $ `G'}
		    \State add `G' to $list\_labels\_sent$
		\ElsIf {$NameOfWordAlignment(l_{k_i})$ is `Insertion' OR `Substitution'}
			 \State add `$B\_ASR$' to $list\_labels\_sent$		
		\Else
			\State add `$B\_MT$' to $list\_labels\_sent$
		\EndIf		
	\EndFor

	\State add $list\_labels\_sent$ to $list\_labels\_result$
\EndFor

\end{algorithmic}
\end{algorithm}

\subsection{Example with 3-label Setting}

Table \ref{tab:example_word_alignment} gives the edit distance between a SLT and MT hypothesis while table \ref{tab:example_3labels}  shows how \textit{Method 1}  and \textit{Method 2} set 3-class labels to the SLT hypothesis. One transcript ($f_{hyp}$) has 1 error. This drives 3 B labels on SLT output ($e_{hyp_{slt}}$), while $e_{hyp_{mt}}$ has only 2 B labels. As can be seen in the cases of \textit{Method 1} and \textit{Method 2}, we respectively have (1 B\_ASR, 2 B\_MT) and (2 B\_ASR, 1 B\_MT). 

\begin{table}[!ht]
\centering
\begin{tabular}{llllllll}
\hline\noalign{\smallskip}
$e_{hyp_{slt}}$ & surgeons & in & los & angeles & it & is & said \\
\noalign{\smallskip}\hline\noalign{\smallskip}

$e_{hyp_{mt}}$  & surgeons & in & los & angeles & ** & have & said \\
\noalign{\smallskip}\hline\noalign{\smallskip}

\multicolumn{1}{c}{\begin{tabular}[c]{@{}c@{}}edit op.\end{tabular}} 
                & Exact & Exact & Exact & Exact & Insertion & Substitution & Exact \\
\noalign{\smallskip}\hline


%
%
%
\end{tabular}
\caption{\label{tab:example_word_alignment}Example of edit distance between SLT and MT.}
\end{table}

\begin{table}[!ht]
\centering
\begin{tabular}{llllllll}
\hline\noalign{\smallskip}
$f_{ref}$ & les chirurgiens & de & los & angeles & ont & & dit \\
\noalign{\smallskip}\hline\noalign{\smallskip}

$f_{hyp}$  & les chirurgiens & de & los & angeles & on & & dit \\
labels ASR & G\space\space G & G & G & G & B & & G \\
\noalign{\smallskip}\hline\noalign{\smallskip}

$e_{hyp_{mt}}$ & surgeons & in & los & angeles &  & have & said \\
labels MT & G & B & G & G &  & B & G \\
\noalign{\smallskip}\hline\noalign{\smallskip}

$e_{hyp_{slt}}$ & surgeons & in & los & angeles & it & is & said \\
labels SLT (2-label)& G & B & G & G & B & B & G \\
labels SLT (\textit{Method 1}) & G & B\_MT & G & G & B\_ASR & B\_MT & G \\
labels SLT (\textit{Method 2}) & G & B\_MT & G & G & B\_ASR & B\_ASR & G \\
\noalign{\smallskip}\hline\noalign{\smallskip}

$e_{ref}$ & the surgeons & of & los & angeles & \space & & said \\
\noalign{\smallskip}\hline


%
%
%
%
%


\end{tabular}

\caption{\label{tab:example_3labels}Example of quintuplet with 2-label and 3-label.}
\end{table}

These differences 
are due to slightly different algorithms for label extraction. As Table \ref{tab:example_3labels} presents, ``is'' (SLT hypothesis) is aligned to ``have'' (MT hypothesis) and ``have'' (MT hypothesis) is labeled by ``B''. It can therefore be assumed that ``is'' (SLT hypothesis) should be annotated with word-level labels by B\_MT 
according to
\textit{Method 1}. However, using \textit{Method 2}, ``is'' (SLT hypothesis) could be labeled by B\_ASR because the type of word alignment between ``is'' (SLT hypothesis) and ``have'' (MT hypothesis) is substitution (S), as shown in Table \ref{tab:example_word_alignment}.


\subsection{Statistics with 3-label Setting on the Whole Corpus}

Table \ref{tab:performances_3labels_slt1_and_slt2} presents the summary statistics for the distribution of $good$ (G), \textit{asr-error} (B\_ASR) and \textit{mt-error} (B\_MT) labels obtained with both label extraction methods.
We see that both methods give similar statistics but slightly different rates of B\_ASR and B\_MT.

\begin{table}[!ht]
\centering
\begin{tabular}{l|rrr|rrr}
\hline

\multicolumn{1}{c|}{\multirow{2}{*}{\textbf{Task - \textit{ASR1}}}} & \multicolumn{3}{c|}{\textbf{\textit{dev} set}}                                                                                              & \multicolumn{3}{c}{\textbf{\textit{tst} set}}                                                                   \\
\multicolumn{1}{c|}{}                               & \multicolumn{1}{c}{\textit{\textbf{\%G}}} & \multicolumn{1}{c}{\textit{\textbf{\%B\_ASR}}} & \multicolumn{1}{c|}{\textit{\textbf{\%B\_MT}}} & \multicolumn{1}{c}{\textbf{\%G}} & \multicolumn{1}{c}{\textbf{\%B\_ASR}} & \multicolumn{1}{c}{\textbf{\%B\_MT}} \\

\hline
label/m1:Method 1 & \textit{62.03} & \textit{19.09} & \textit{18.89} & 70.59 & 14.50  & 14.91 \\
label/m2:Method 2 & \textit{62.03} & \textit{22.49} & \textit{15.49} & 70.59 & 16.62  & 12.79 \\
label/same(m1, m2)          & \textit{62.03} & \textit{18.09} & \textit{14.49} & 70.59 & 13.58  & 11.88 \\
label/diff(m1, m2)          & \textit{0}     & \textit{ 1.00} & \textit{ 4.40}  & 0     &  0.92  &  3.03  \\

\hline
\hline

\multicolumn{1}{c|}{\multirow{2}{*}{\textbf{Task - \textit{ASR2}}}} & \multicolumn{3}{c|}{\textbf{\textit{dev} set}}                                                                                              & \multicolumn{3}{c}{\textbf{\textit{tst} set}}                                                                   \\
\multicolumn{1}{c|}{}                               & \multicolumn{1}{c}{\textit{\textbf{\%G}}} & \multicolumn{1}{c}{\textit{\textbf{\%B\_ASR}}} & \multicolumn{1}{c|}{\textit{\textbf{\%B\_MT}}} & \multicolumn{1}{c}{\textbf{\%G}} & \multicolumn{1}{c}{\textbf{\%B\_ASR}} & \multicolumn{1}{c}{\textbf{\%B\_MT}} \\

\hline
label/m1:Method 1 & \textit{63.87} & \textit{16.89} & \textit{19.23} & 72.61 & 11.92 & 15.47 \\
label/m2:Method 2 & \textit{63.87} & \textit{19.78} & \textit{16.34} & 72.61 & 13.58 & 13.81 \\
label/same(m1, m2)          & \textit{63.87} & \textit{16.05} & \textit{15.50} & 72.61 & 11.12 & 13.01 \\
label/diff(m1, m2)          & \textit{0}     & \textit{ 0.84} & \textit{ 3.73} & 0     &  0.80 &  2.46 \\ 

\hline

\end{tabular}
\caption{\label{tab:performances_3labels_slt1_and_slt2}Statistics with 3-label setting for \textit{ASR1} and \textit{ASR2}.}
\end{table}

As can be seen from Table \ref{tab:performances_3labels_slt1_and_slt2}, it is interesting to note that while ASR system improves from \textit{ASR1} to \textit{ASR2}, the rate of B\_ASR labels logically decreases by more than 2 points, while the rate of B\_MT remains almost stable (less than 1 point difference) which makes sense since the MT system is the same in
both \textit{ASR1} and \textit{ASR2}.
These statistics show that intersection between both methods is probably a good estimation of disentangled ASR and MT errors in SLT.


\subsection{Qualitative Analysis of SLT Errors}

Our new 3-label setting procedure allows us to analyze the behavior of our SLT system. 


\begin{table}[!ht]
\centering
\begin{tabular}{p{0.8cm}p{11cm}}
\hline\noalign{\smallskip}

\textit{$f_{ref}$} & peter frey est né le quatre août mille neuf cent cinquante sept à bingen \\
\textit{$f_{hyp_{1}}$} & \textbf{pierre ferait aimé} le quatre août mille neuf cent cinquante sept à \textbf{big m} \\
\textit{$f_{hyp_{2}}$} & \textbf{pierre} frey est né le quatre août mille neuf cent cinquante sept à \textbf{big m} \\
\textit{$e_{hyp_{mt}}$} & peter frey was born on \textbf{4} august 1957 \textbf{to} bingen . \\
\textit{$e_{hyp_{slt1}}$} & \textbf{pierre would liked the four} august \textbf{thousand nine hundred and fifty seven to big m} \\
\textit{$e_{hyp_{slt2}}$} & \textbf{pierre} frey is born \textbf{the four} august \textbf{thousand nine hundred and fifty seven to big m} \\
\textit{$e_{ref}$} & peter frey was born on august 4th 1957 in bingen . \\

\noalign{\smallskip}\hline
\end{tabular}
\caption{\label{tab:examples_3labels_1}Example 1 - SLT hypothesis annotated with two methods - having a few \textit{asr-errors}, a few \textit{mt-errors} and many \textit{slt-errors} such as 5 B\_ASR1, 3 B\_ASR2, 2 B\_MT, 14 B\_SLT1, 12 B\_SLT2.}
\end{table}

We can observe sentences with Table \ref{tab:examples_3labels_1} presents, as an example, 
few ASR and MT errors leading to many SLT errors. Indeed, this is a good way of detecting flaws in the SLT pipeline such as bad post-processing of the SLT output (numerical or text dates, for instance).

As shown in Table \ref{tab:examples_3labels_2}, on the contrary, there are many ASR errors leading to few SLT errors (ASR errors with few consequences such as morphological substitutions - for instance in French: de/des, déficit/déficits, budgétaire/budgétaires).

Finally, ASR errors as presented in Table \ref{tab:examples_3labels_3} have different consequences on SLT quality (on a sample sentence, 2 ASR errors of system 1 and 2 lead to 14 and 9 SLT errors, respectively).

\begin{figure}[!ht]
  \centering
  \begin{minipage}[b]{\textwidth}
    \includegraphics[width=\textwidth]{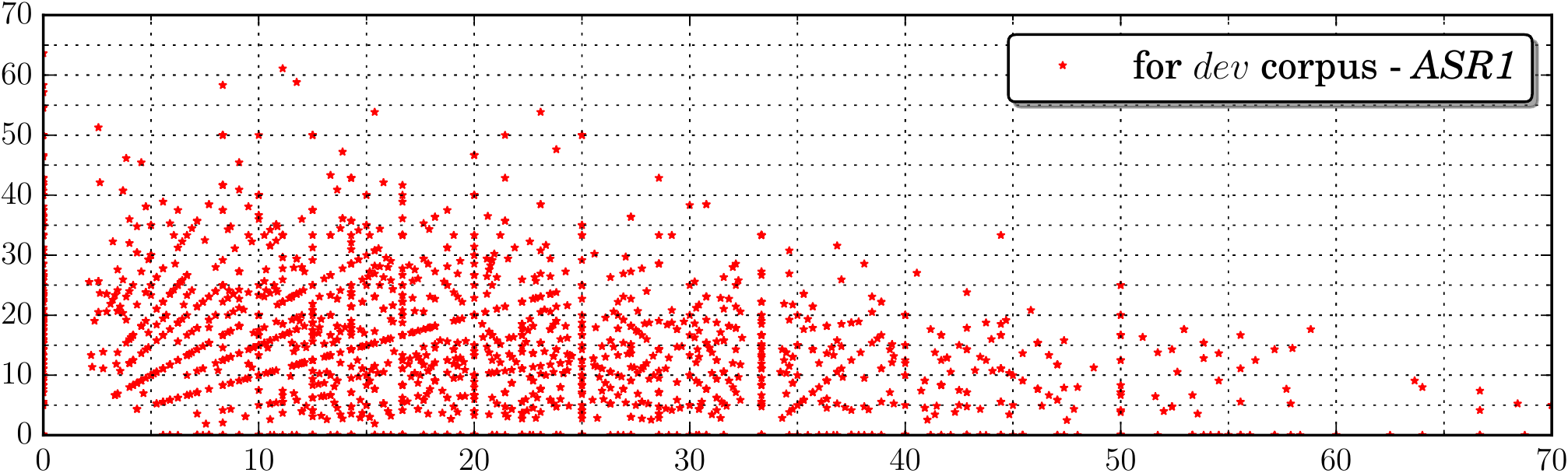}
  \end{minipage}
  \hfill
%
  \begin{minipage}[b]{\textwidth}
    \includegraphics[width=\textwidth]{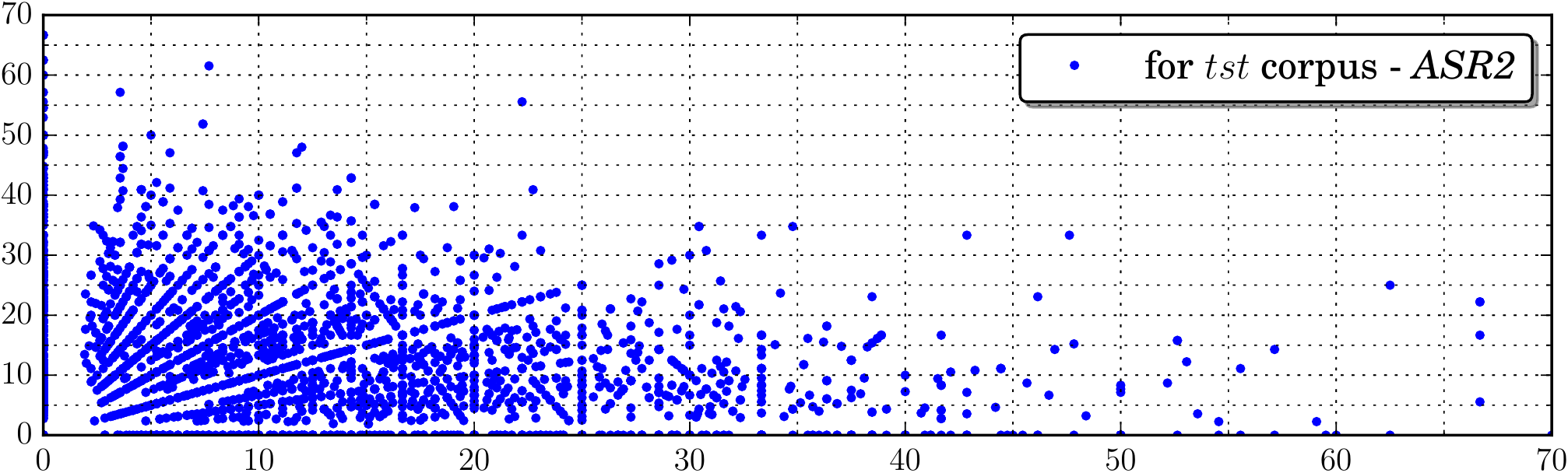}
  \end{minipage}

\caption{\label{fig:figure_percentage_for_intersection_ASR1_and_ASR2} Example of the rate (\%) of ASR errors (x-axis) versus (\%) MT errors (y-axis) - for \textit{dev/ASR1} and \textit{tst/ASR2}.}
\end{figure}

Figure \ref{fig:figure_percentage_for_intersection_ASR1_and_ASR2} shows how our speech utterances are distributed in the two-dimensional ($B_{ASR}$, $B_{MT}$) error space.





\begin{table}[!ht]
\centering
\begin{tabular}{p{0.8cm}p{11cm}}
\hline\noalign{\smallskip}

\textit{$f_{ref}$} & malheureusement le système européen de financement gouvernemental direct est \\
\textit{$f_{hyp_{1}}$} & malheureusement le système européen financement \textbf{gouvernementale directe et} \\
\textit{$f_{hyp_{2}}$} & malheureusement le système européen de financement gouvernemental direct est \\
\textit{$e_{hyp_{mt}}$} & unfortunately , the european system of direct government funding is \\
\textit{$e_{hyp_{slt1}}$} & unfortunately the european system direct government funding \\
\textit{$e_{hyp_{slt2}}$} & unfortunately the european system of direct government funding is \\
\textit{$e_{ref}$} & unfortunately , the european system of direct government funding is \\

\hline\noalign{\smallskip}
\hline\noalign{\smallskip}

\textit{$f_{ref}$} & victime de la croissance économique européenne lente et des déficits budgétaires \\
\textit{$f_{hyp_{1}}$} & \textbf{victimes} de la croissance économique européenne \textbf{venant de déficit budgétaire} \\
\textit{$f_{hyp_{2}}$} &  victime de la croissance économique européenne \textbf{venant} des déficits budgétaires \\
\textit{$e_{hyp_{mt}}$} & a victim of european economic growth \textbf{slow} and budget deficits . \\
\textit{$e_{hyp_{slt1}}$} & \textbf{and} victims of european economic growth \textbf{from} budget deficit \\
\textit{$e_{hyp_{slt2}}$} & a victim of european economic growth \textbf{from the} budget deficits \\
\textit{$e_{ref}$} & a victim of slow european economic growth and budget deficits . \\

\noalign{\smallskip}\hline
\end{tabular}
\caption{\label{tab:examples_3labels_2}Example 2 - SLT hypothesis annotated with two methods - having many \textit{asr-errors}, a few \textit{mt-errors} and a few \textit{slt-errors} such as 8 B\_ASR1, 1 B\_ASR2, 1 B\_MT, 2 B\_SLT1, 2 B\_SLT2.}
\end{table}


\begin{table}[!ht]
\centering
\begin{tabular}{p{0.8cm}p{11cm}}
\hline\noalign{\smallskip}

\textit{$f_{ref}$} & nous ne comprenons pas ce qui se passe chez les jeunes pour qu' ils trouvent \\
\textit{$f_{hyp_{1}}$} & nous ne comprenons pas \textbf{ceux} qui se passe chez les jeunes pour qu' ils trouvent \\
\textit{$f_{hyp_{2}}$} & nous ne comprenons pas ce qui se passe chez les jeunes pour qu' \textbf{il trouve} \\
\textit{$e_{hyp_{mt}}$} & we do not understand what is happening \textbf{among} young people for \textbf{that}  \\
\textit{$e_{hyp_{slt1}}$} & we do not understand \textbf{those who happens among} young people for \textbf{that}   \\
\textit{$e_{hyp_{slt2}}$} & we do not understand what is happening \textbf{among} young people \\
\textit{$e_{ref}$} & we do not understand what is happening in young people 's mind for them \\

\hline\noalign{\smallskip}
\hline\noalign{\smallskip}

\textit{$f_{ref}$} & amusant de maltraiter gratuitement un animal sans défense qui nous donne \\
\textit{$f_{hyp_{1}}$} & amusant de \textbf{maltraité} gratuitement un animal sans défense qui nous  \\
\textit{$f_{hyp_{2}}$} & amusant de maltraiter gratuitement un animal sans défense qui nous donne \\
\textit{$e_{hyp_{mt}}$} & \textbf{they are fun} to mistreat \textbf{free a} defenceless animal  \\
\textit{$e_{hyp_{slt1}}$} & \textbf{they} find \textbf{fun free} mistreated \textbf{a} defenceless animal \\
\textit{$e_{hyp_{slt2}}$} & to find \textbf{it} amusing to mistreat \textbf{free a} defenceless animal\\
\textit{$e_{ref}$} & to find amusing to mistreat defenceless animals without reason ,  \\

\hline\noalign{\smallskip}
\hline\noalign{\smallskip}

\textit{$f_{ref}$} & de l' affection de l' amitié et nous tient compagnie \\
\textit{$f_{hyp_{1}}$} & de l' affection de l' amitié nous tient compagnie \\
\textit{$f_{hyp_{2}}$} & de l' affection de l' amitié nous tient compagnie \\
\textit{$e_{hyp_{mt}}$} & which gives us \textbf{the affection} , friendship and \textbf{keeps us airline} . \\
\textit{$e_{hyp_{slt1}}$} & which \textbf{we affection of} friendship \textbf{we takes} company  \\
\textit{$e_{hyp_{slt2}}$} & which gives us \textbf{the affection of} friendship \textbf{we takes} company  \\
\textit{$e_{ref}$} & which gives us love , friendship and companionship . \\

\noalign{\smallskip}\hline
\end{tabular}
\caption{\label{tab:examples_3labels_3}Example 3 - SLT hypothesis annotated with two methods - having the same number of \textit{asr-errors}, but the different number of \textit{slt-errors} extracted from \textit{ASR1} and \textit{ASR2} such as 2 B\_ASR1, 2 B\_ASR2, 12 B\_MT, 14 B\_SLT1, 9 B\_SLT2.}
\end{table}


\section{Automatic Error Detection for SLT}
\label{sec:WCEjoint}

In this paper, we use Conditional Random Fields \citep{lafferty01} (CRFs) as our machine learning method, with WAPITI toolkit \citep{lavergne10}, to train our error detector
based on MT and ASR engineered features. For ASR, we extract 9 features, which come from the ASR graph, from language model scores  and from a morphosyntactic analysis. These detailed features could be found in 
\citep{besacier14}.
For MT, we use a total of 24 major feature types which can be extracted with our word confidence estimation toolkit for MT (more details are given in 
\citep{servan-toolkit-2015}).

\subsection{Experiments on 2-class Error Detection}

\begin{table}[!ht]
\centering
\begin{tabular}{lll}
\hline\noalign{\smallskip}
Exp         & MT+ASR feat.   &Joint feat. \\
                &     $p_{ASR}(q|x_{f},f)^\alpha$     &  $p(q|x_{f},f,e)$  \\
                &     $*p_{MT}(q|e,f)^{1-\alpha}$     &          \\

\noalign{\smallskip}\hline\noalign{\smallskip}
\textit{F-avg1 (ASR1)} & 58.07\%&64.90\%\\
\textit{F-avg2 (ASR2)} & 53.66\%&64.17\%\\
\noalign{\smallskip}\hline
\end{tabular}
\caption{\label{tab:wce_performance_diff_feat_for_tst_set}Error Detection Performance (2-label) on SLT ouptut for \textit{tst} set  (training is made on \textit{dev} set).}
\end{table}

In this experiment, we evaluate the performance of our classifiers by using the average between the F-measure for $good$ labels and the F-measure for $bad$ labels that are calculated by the common evaluation metrics: Precision, Recall and F-measure for $good$/$bad$ labels. Since two ASR systems are available, \textit{F-avg1} is obtained for SLT based on $ASR1$ whereas \textit{F-avg2} is obtained for SLT based on $ASR2$.  
The classifier is evaluated on the \textit{tst} part of our corpus and trained on the \textit{dev} part. 


We report in Table \ref{tab:wce_performance_diff_feat_for_tst_set} the baseline error detection results obtained using both MT and ASR features for a 2-class problem (error detection). More precisely we evaluate two different approaches (\textit{combination} and \textit{joint}):

\begin{itemize}
\item First system (MT+ASR feat.) combines the output of two separate classifiers based on ASR and MT features. In this approach, ASR-based confidence score of the source is projected to the target SLT output and combined with the MT-based confidence score as shown in
Equation \ref{eq:wce_asr_and_mt}
(we did not tune the $\alpha$ coefficient and set it \textit{a priori} to 0.5).

\item Second system (joint feat.) trains a single error detection system for SLT (evaluating $p(q|x_{f},f,e)$ as in 
Equation \ref{eq:wce_component} 
using joint ASR and MT features. ASR features are projected to the target words using automatic word alignments.

\end{itemize}

Table \ref{tab:wce_performance_diff_feat_for_tst_set} shows that joint ASR and MT features improve error detection performance over the use of simple combination (MT+ASR). Based on this result, only the joint approach is used in our 3-class experiments of next section.
We also observe that F-measure decreases when ASR WER is lower (\textit{F-avg2}$<$\textit{F-avg1} while $WER_{ASR2}<WER_{ASR1}$). So error detection for SLT might be more complicated as ASR system improves. 

These observations lead us to investigate the behaviour of our WCE approaches for a large range of $good$/$bad$ decision threshold.

\begin{figure}[!tbp]
  \centering
  \begin{minipage}[b]{0.7\textwidth}
    \includegraphics[width=\textwidth]{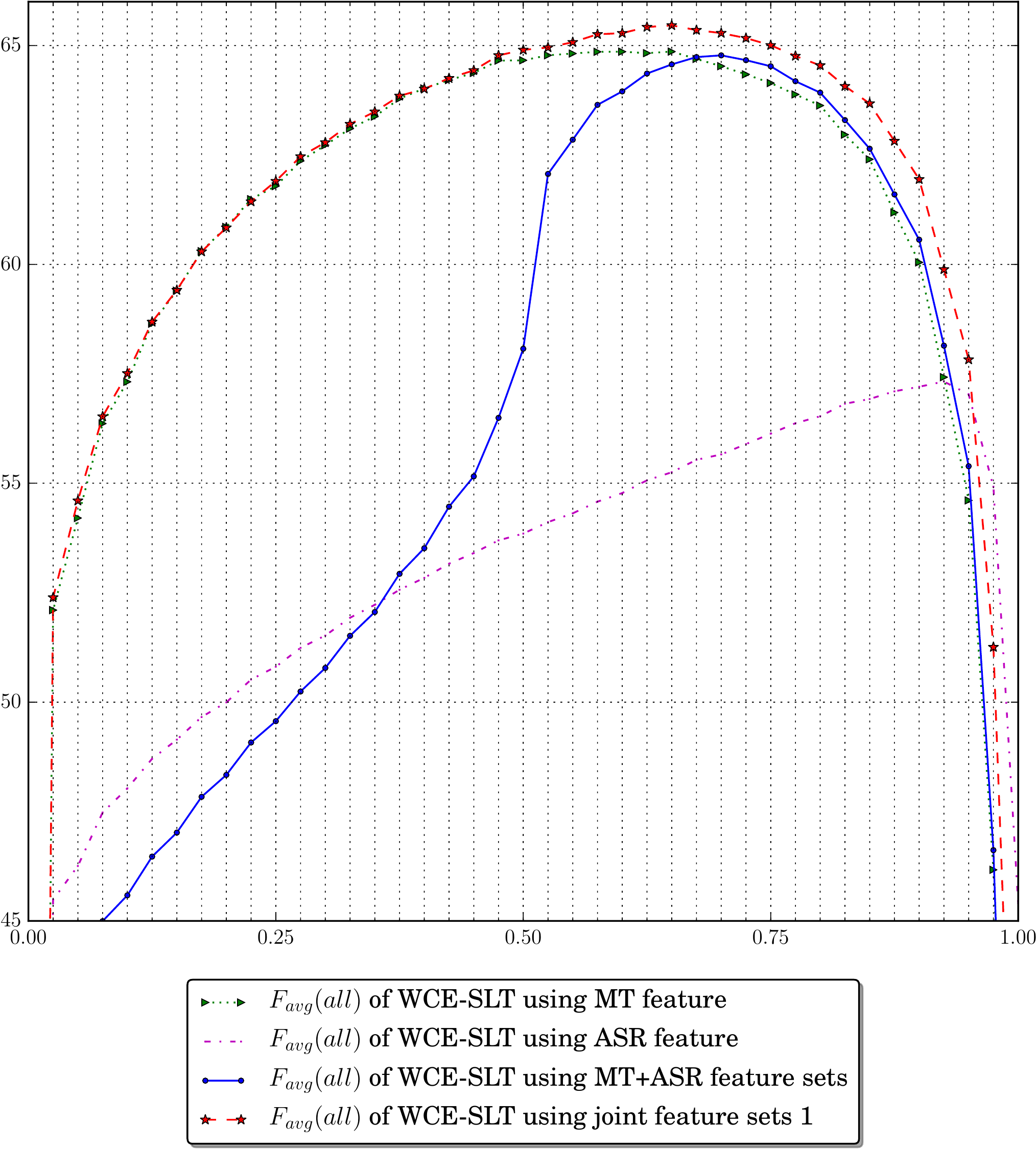}
    
  \end{minipage}
  \hfill
  \begin{minipage}[b]{0.7\textwidth}
    \includegraphics[width=\textwidth]{curves_TestTST-F_mes_avg_all-tgt-slt1-after-terpa.pdf}    
  \end{minipage}
  \caption{\label{fig:WCE_SLT_for_MT_ASR_feat_AND_Joint_feat_for_TestTST_corpus_all_curves1_and_2}Evolution of system performance (y-axis - \textit{F-mes1} - ASR1 and \textit{F-mes2} - ASR2) for \textit{tst} corpus (4050 utt) along decision threshold variation (x-axis) - training is made on \textit{dev} corpus (2643 utt).}
\end{figure}


While the previous tables provided WCE performance for a single point of interest ($good$/$bad$ decision threshold set to 0.5), the curves of Figure 
\ref{fig:WCE_SLT_for_MT_ASR_feat_AND_Joint_feat_for_TestTST_corpus_all_curves1_and_2}
show the full picture of our WCE systems (for SLT) using speech transcriptions systems $ASR1$ and $ASR2$, respectively. We observe that the classifier based on ASR features has a very different behaviour than the classifier based on MT features which explains why their simple combination (MT+ASR) does not work very well for the default decision threshold (0.5). However, for threshold above 0.75, the use of both ASR and MT features is slightly beneficial. This is interesting because higher thresholds improves the F-measure on $bad$ labels (so improves error detection). Both curves are similar whatever the ASR system used. These results suggest that with enough development data for appropriate threshold tuning (which we do not have for this very new task), the use of both ASR and MT features should improve error detection in speech translation (blue and red curves are above the green curve for higher decision threshold\footnote{Corresponding to optimization of the F-measure on $bad$ labels (errors).}). 

\subsection{Experiments on 3-class Error Detection}
We report in Table \ref{tab:wce_performance_joint_feat_for_tst_set_for_ASR1_and_ASR2_detail}  our first attempt to build an error detection system in SLT as a 3-class problem (\textit{joint} approach only). 
We made our experiment by training and evaluating the model on \textit{Intersection(m1, m2)} which corresponds to high confidence in the labels\footnote{However, we observed (results not reported here) that the use of different label sets (\textit{Method 1}, \textit{Method 2}, \textit{Intersection(Method 1, Method 2)} does not have a strong influence on the results.}. 
We compared two different approaches: \textit{One-Step} is a single classifier for the 3-class problem while
\textit{Two-Step} first applies the 2 class (\textit{G/B}) system and a second classifier distinguishes $B_{ASR}$ and $B_{MT}$ errors. Not much difference in F-measure is observed between both approaches. 
Table \ref{tab:stat_on_correctly_detected_bad_on_ASR1_and_ASR2} also presents the confusion matrix between $B_{ASR}$ and $B_{MT}$ for the correctly detected (true) errors. Despite the relatively low F-scores of table \ref{tab:wce_performance_joint_feat_for_tst_set_for_ASR1_and_ASR2_detail}, we see that our 3-labels classifier obtains encouraging confusion matrices in order to automatically disentangle $B_{ASR}$ and $B_{MT}$ on true errors.

\begin{table}[!ht]
\centering
\begin{tabular}{lll|lllll}
\hline


\bf    & \multicolumn{2}{c|}{\multirow{2}{*}{\textbf{2-class}}} &  & \multicolumn{4}{c}{\multirow{2}{*}{\textbf{3-class}}}   \\

\bf    & \multicolumn{2}{c|}{\multirow{2}{*}{\textbf{Full Corpus}}} &  & \multicolumn{4}{c}{\multirow{2}{*}{\textbf{Intersection Corpus (m1, m2)}}}   \\

\bf    &    & &   \\
\hline           

		   &  &  &  &  \multicolumn{2}{c}{\textbf{One-Step}}   &  \multicolumn{2}{c}{\textbf{Two-Step}}  \\

           & \textit{\textbf{ASR1}} & \textit{\textbf{ASR2}}   &  &  \textit{\textbf{ASR1}} & \textit{\textbf{ASR2}}   &  \textit{\textbf{ASR1}} & \textit{\textbf{ASR2}}  \\

\hline
$F_G$        & 81.79 & 83.17 & $F_G$         & 85.00 & 85.00 & 84.00 & 85.00 \\
$F_B$        & 48.00 & 45.17 & $F_{B\_ASR}$  & 44.00 & 42.00 & 44.00 & 42.00 \\
		  &	      &          & $F_{B\_MT}$   & 14.00 & 15.00 & 16.00 & 17.00 \\
$F_{avg}$    & 64.90 & 64.17 & $F_{avg}$     & 47.67 & 47.33 & 48.00 & 48.00 \\
\hline
\end{tabular}
\caption{\label{tab:wce_performance_joint_feat_for_tst_set_for_ASR1_and_ASR2_detail}Error Detection Performance (2-label vs 3-label) on SLT output for tst set (training is made on \textit{dev} set).}
\end{table}


\begin{table}[!ht]
\centering
\begin{tabular}{l|rr|rr}
\hline

\multicolumn{1}{c|}{\multirow{2}{*}{\textbf{(1) Ref $\backslash$ Hyp}}} & \multicolumn{2}{c|}{\textbf{\textit{ASR1}}}                                                                                              & \multicolumn{2}{c}{\textbf{\textit{ASR2}}}                                                                   \\
\multicolumn{1}{c|}{}  & \multicolumn{1}{c}{\textit{\textbf{B\_ASR}}} & \multicolumn{1}{c|}{\textit{\textbf{B\_MT}}} & \multicolumn{1}{c}{\textbf{B\_ASR}} & \multicolumn{1}{c}{\textbf{B\_MT}} \\

\hline
\textit{\textbf{B\_ASR}} & \textit{85.75\%} & \textit{14.25\%} & 81.57\% & 18.43\% \\
\textit{\textbf{B\_MT}} & \textit{44.46\%} & \textit{55.54\%} & 34.53\% & 65.47\% \\

\hline
\hline

\multicolumn{1}{c|}{\multirow{2}{*}{\textbf{(2) Ref $\backslash$ Hyp}}} & \multicolumn{2}{c|}{\textbf{\textit{ASR1}}}                                                                                              & \multicolumn{2}{c}{\textbf{\textit{ASR2}}}                                                                   \\
\multicolumn{1}{c|}{}  & \multicolumn{1}{c}{\textit{\textbf{B\_ASR}}} & \multicolumn{1}{c|}{\textit{\textbf{B\_MT}}} & \multicolumn{1}{c}{\textbf{B\_ASR}} & \multicolumn{1}{c}{\textbf{B\_MT}} \\

\hline
\textit{\textbf{B\_ASR}} & \textit{83.14\%} & \textit{16.86\%} & 80.02\% & 19.98\% \\
\textit{\textbf{B\_MT}} & \textit{49.41\%} & \textit{50.59\%} & 41.49\% & 58.51\% \\

\hline

\end{tabular}
\caption{\label{tab:stat_on_correctly_detected_bad_on_ASR1_and_ASR2}Confusion Matrix on Correctly Detected Errors Subset for 3-class (1) One-Step; (2) Two-Step.}
\end{table}


%
%
%

\section{Conclusions}
\label{sec:illust}

This paper proposed to disentangle ASR and MT errors in speech translation. The binary error detection problem was recast as a 3-class labeling problem (\textit{good, asr-error, mt-error}). First, two methods were proposed for the non trivial label setting and it was shown that both give consistent results. Then, automatic detection of error types, using joint ASR and MT features, was evaluated and encouraging results were displayed on a French-English speech translation task. We believe that such a new task (not only detecting errors but also their cause) is interesting to build better informed speech translation systems, especially in interactive speech translation use cases.

\small

\bibliographystyle{apalike}

\bibliography{interspeech2017}


\end{document}